\documentclass[letterpaper, 10 pt, conference]{ieeeconf}  

\IEEEoverridecommandlockouts                              

\usepackage[hidelinks]{hyperref}

\usepackage{graphicx} 
\usepackage{amsmath} 
\usepackage{amssymb}  
\usepackage{algorithm,algorithmicx} 
\usepackage{algpseudocode}
\usepackage{cleveref} 
\usepackage{xspace} 
\usepackage{xcolor}
\usepackage[percent]{overpic} 
\usepackage{caption}
\usepackage{subcaption} 
\usepackage{tablefootnote} 
\usepackage{makecell} 

\usepackage{flushend}

\newcommand{\MS}{Motion~Stack\xspace}
\newcommand{\MB}{MoonBot\xspace}
\newcommand{\MBs}{MoonBots\xspace}
\newcommand{\MST}{MoonShot\xspace}

\newcommand{\ros}{ROS2\xspace}

\newcommand{\PY}{Python\xspace}

\crefname{section}{Sec.}{Secs.} 
\crefname{figure}{Fig.}{Figs.} 
\crefname{equation}{}{}
\crefname{algorithm}{Alg.}{Algs.}
\crefname{table}{Tab.}{Tabs.}

\crefname{section}{Sec.}{Secs.}
\crefname{figure}{Fig.}{Figs.}
\crefname{equation}{}{}
\crefname{algorithm}{Alg.}{Algs.}
\crefname{table}{Tab.}{Tabs.}

\title{\LARGE \bf
Designing for Distributed Heterogeneous Modularity: \\On Software Architecture and Deployment of the \MBs
}

\author{Elian Neppel$^{1}$, Shamistan Karimov$^{1}$, Ashutosh Mishra$^{1}$, Gustavo H. Diaz$^{1}$, Hazal Gozbasi$^{1}$, \\Shreya Santra$^{1}$, Kentaro Uno$^{1}$ and Kazuya Yoshida$^{1}$
\thanks{
$^{*}$This work was supported by JST SPRING, Grant Number JPMJSP2114, and JST Moonshot R\&D Program, Grant Number JPMJMS223B.
    }%
\thanks{$^{1}$All authors are with the Space Robotics Lab. (SRL) in Department of Aerospace Engineering, Graduate School of Engineering, Tohoku University, Sendai 980--8579, Japan. (E-mail: \tt{neppel.elian.s6@dc.tohoku.ac.jp})  }%
\thanks{
\textit{The corresponding author is Elian Neppel.}
    }
}

\begin{document}

\maketitle
\thispagestyle{empty}
\pagestyle{empty}

\begin{abstract}

This paper presents the software architecture and deployment strategy behind the \MB platform: a modular space robotic system composed of heterogeneous components distributed across multiple computers, networks and ultimately celestial bodies. We introduce a principled approach to distributed, heterogeneous modularity, extending modular robotics beyond physical reconfiguration to software, communication and orchestration.
We detail the architecture of our system that integrates component-based design, a data-oriented communication model using \ros and Zenoh, and a deployment orchestrator capable of managing complex multi-module assemblies. These abstractions enable dynamic reconfiguration, decentralized control, and seamless collaboration between numerous operators and modules. At the heart of this system lies our open-source \MS software, validated by months of field deployment with self-assembling robots, inter-robot cooperation, and remote operation.
Our architecture tackles the significant hurdles of modular robotics by significantly reducing integration and maintenance overhead, while remaining scalable and robust. Although developed with space in mind, we propose generalizable patterns for designing robotic systems that must scale across time, hardware, teams and operational environments.

\end{abstract}


\section{Introduction}

\label{sec:intro}

Achieving true modularity in robotics would mark a significant leap
forward~\cite{survey_reconf, Yim2009}, especially in the context of space
exploration~\cite{in_space_assembly, space_reconf2021, colab2020}. Most robotic
systems today are engineered as monolithic, task-specific platforms, requiring
the task or environment to adapt to the robot rather than the
opposite~\cite{robust_cheeta2018,saber2021,quatro2014}. This rigid coupling
between form and function significantly limits adaptability.

\begin{figure}[tb]
    \centering

    \begin{overpic}[width=1\columnwidth, trim={100 230 130 69}, clip]{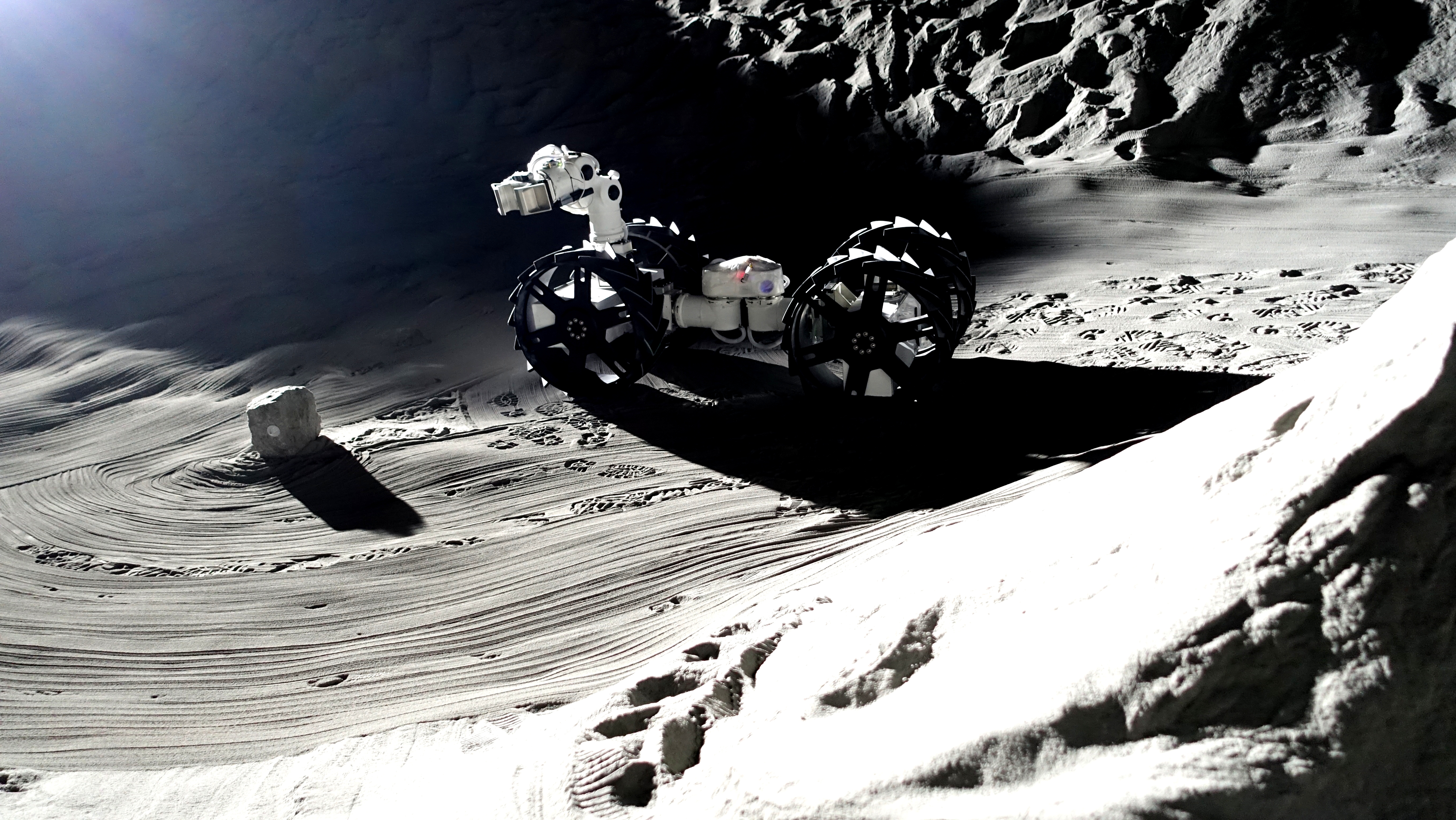}
        \put(2,33){\color{white}\textbf{(a)}}
    \end{overpic}
    \par\vskip5pt
    \begin{overpic}[width=1\columnwidth, trim={100 130 0 80}, clip]{"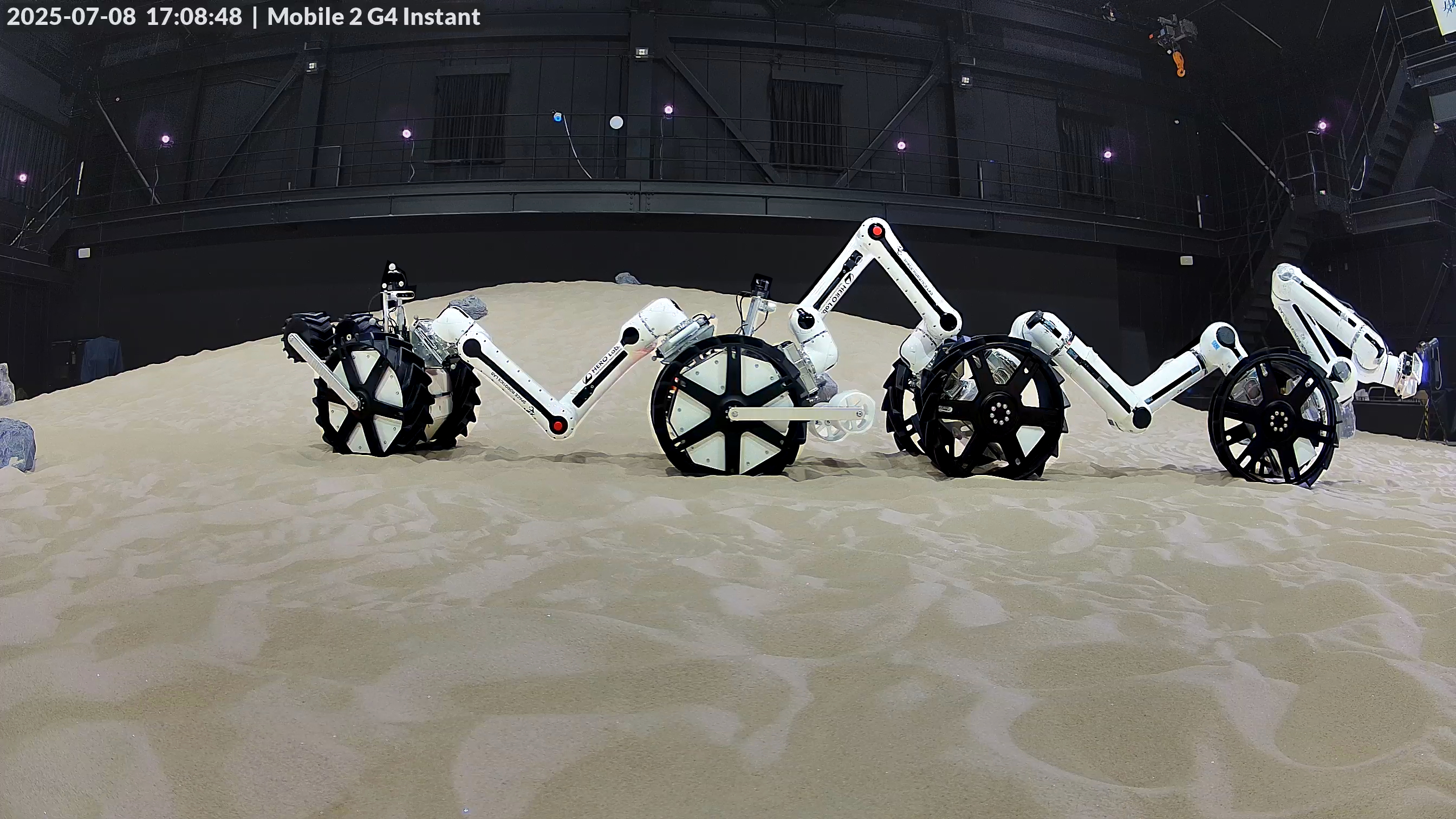"}
        \put(2,23){\color{white}\textbf{(b)}}
    \end{overpic}
    \par\vskip3pt
    \begin{overpic}[height=0.327\columnwidth, trim={180 0 130 130}, clip]{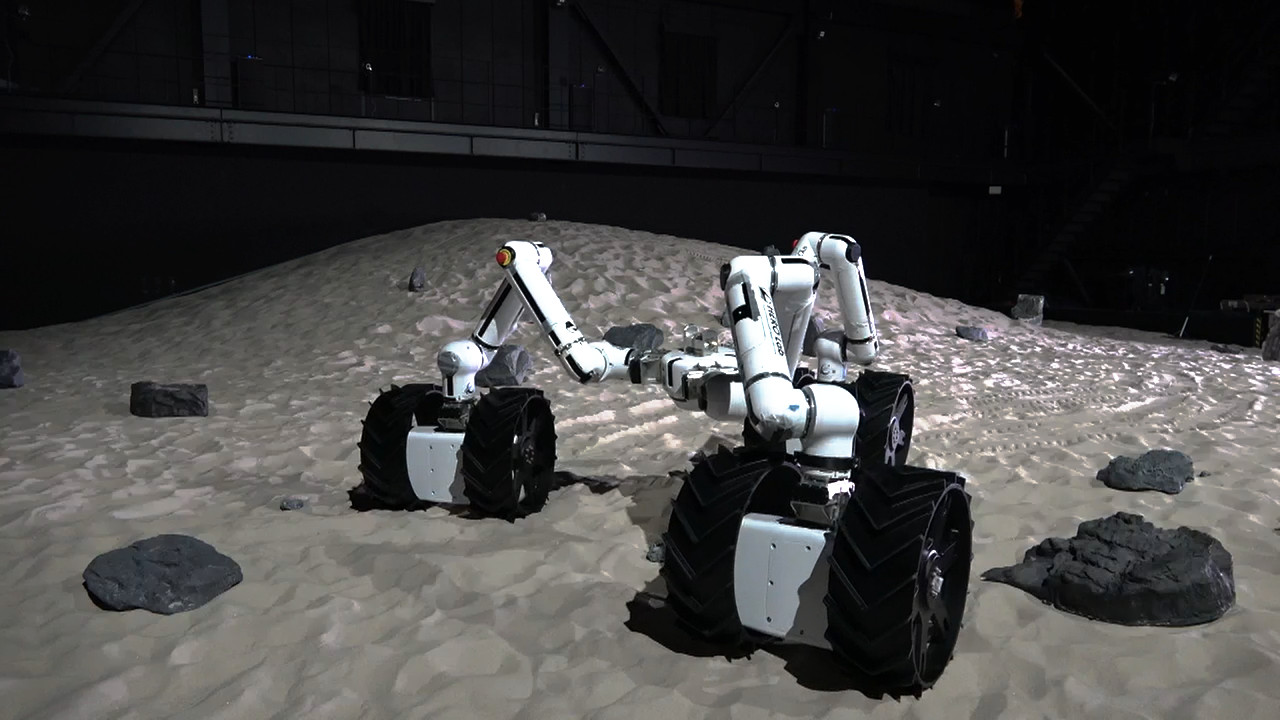}
        \put(3,55){\color{white}\textbf{(c)}}
    \end{overpic}
    \begin{overpic}[height=0.327\columnwidth, trim={120 30 100 30}, clip]{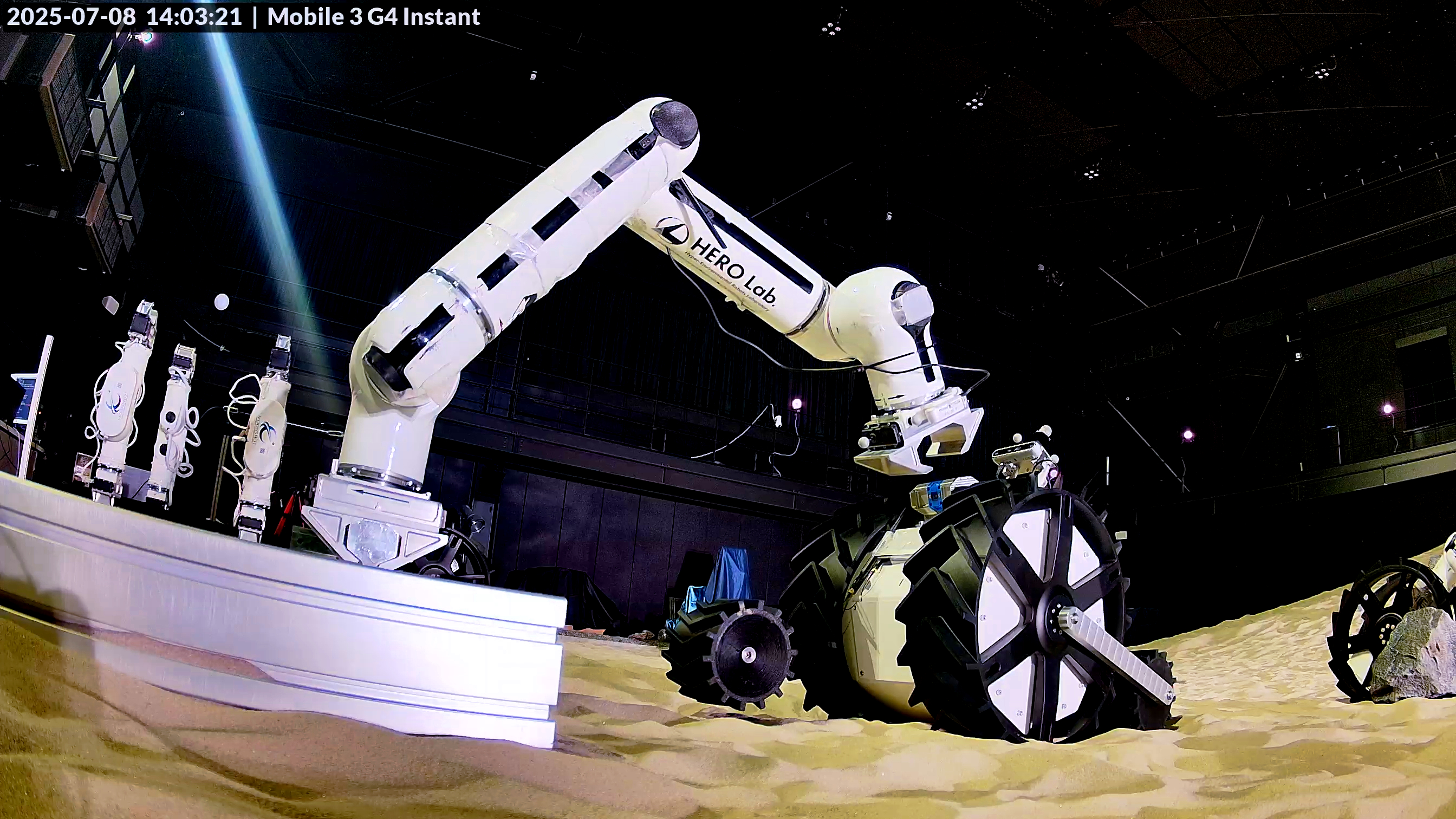}
        \put(3,61){\color{white}\textbf{(d)}}
    \end{overpic}

    \caption{A variety of \MB assemblies. (a):~Dragon assembly -- 4 modules, made of 2 \MB G-line limbs and 2 \MB H-line V1 wheels. Specialized in nimble navigation. (b):~Dragon x4 assembly -- 8 modules, made of a mix of \MB H-line V1 and V2, wheels and limbs. Specialized in redundancy. (c):~Tricycle assembly -- 6~\MB H-line V1 modules. Spacialized in cargo. (d):~\MB H-line V1 limb and V2 wheel assembling themselves.}
    \label{fig:mainx4}
\end{figure}

As a concrete example, no single robot can palletize boxes on an assembly line,
drive on public roads, and traverse mountainous terrain. While excellent
technologies exist for each task individually~\cite{MoveIt2019, self_driving,
robust_cheeta2018}, they remain siloed and cannot be unified into a single,
adaptive system. Humans, in contrast, excel at switching between domains -- a
capability that continues to inspire the pursuit of general-purpose humanoid
robots~\cite{humanoid}. But, as illustrated in \cref{fig:mainx4}, modular
robots offer an alternative vision: breaking the monolith to build
reconfigurable systems capable of adapting to the task, environment, and
mission~\cite{space_reconf2021, survey_reconf, Yim2009}.

\begin{figure}[tb]
    \vspace{1.8mm}
	\centering
	\includegraphics[width=1\columnwidth]{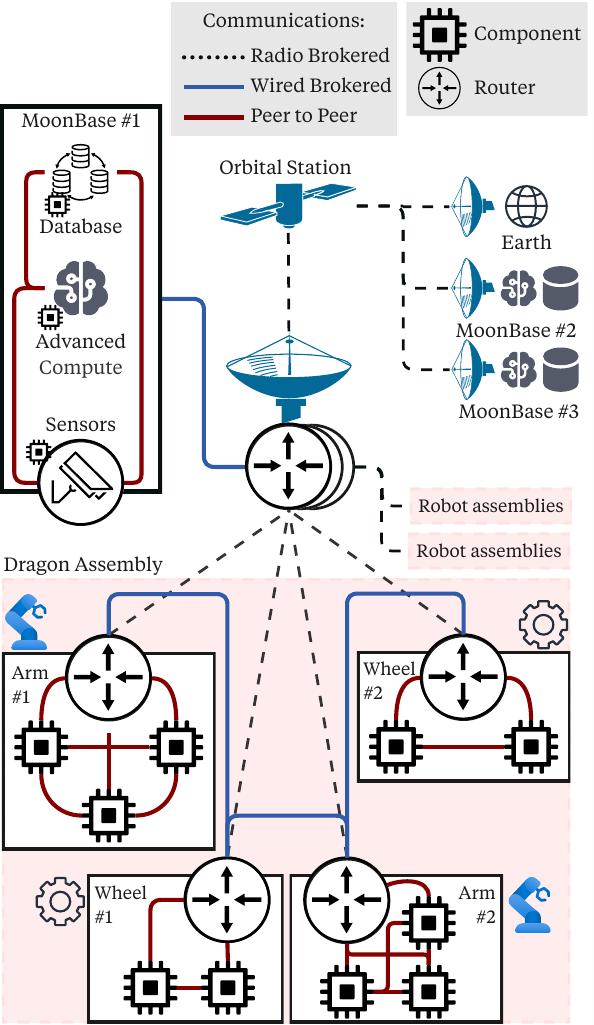}
    \caption{Zenoh network overview of distributed Moon bases and robot
    assemblies. It combines multiple topologies to maximize reliability and
    performance under various communication interfaces.}
	\label{fig:interplanet}
\end{figure}

Our project: \textit{\MST} \cite{moonbot_webpage}, envisions a fleet of \MB robots building lunar infrastructures. As illustrated 
\cref{fig:mainx4,fig:interplanet} and described in our article~\cite{moonbot},
\MBs reconfigure in a multitude of shapes, specialized in diverse tasks --
exploration, mapping, construction, and repair -- and distributed over wide
networks. Modularity is key for such space projects~\cite{colab2020,
space_busi_2020}: Redundancy is improved with replaceable parts; weight is
reduced by sending modules instead of specialized robots; load is balanced
between servers; the system can be upgraded through the
years~\cite{in_space_assembly, space_reconf2021}.

To proceed, we must clarify our broader notion of \textit{modularity}. While
existing robotic frameworks define it through physical
reconfiguration~\cite{moobot_zero, ModularReconfig2020}, our approach
distinguishes itself by not tying modularity to physical shape. In the
literature we find three emergent family of modularity: ``graft modularity''
(augmenting capability by adding components~\cite{saber2021, quatro2014}),
``clone modularity'' (scaling via identical units~\cite{Yim2009,
decentra_robot_mod}), and ``swarm modularity'' (emergent behavior from many
similar agents~\cite{self_driving, swarm}). Our framework aim for a superset,
novel to robotics, encompassing all those approaches: ``distributed
heterogeneous modularity,'' where modules differ in form, function, and
communication, and are distributed across both physical and computational
infrastructure.

We can find our large-scale modularity practiced in space
systems~\cite{in_space_assembly}. The International Space Station 
exemplifies this mindset with heterogeneous physical modules, but also distributed
components across orbit and ground, cooperating for control, analysis, and
fault recovery. 

Distributed heterogeneous modularity is now widespread in modern software, with
Internet of Tings (IoT) and cloud systems~\cite{iiot}. These domains
have developed mature, scalable patterns for managing complexity across
heterogeneous layers of infrastructure. We observe those technologies -- DDS,
MQTT, and Zenoh~\cite{DDS, MQTT, zenoh} -- arriving in the field of robotics,
giving birth to Industrial IoT (IIoT)~\cite{iiot}. Based on
those ideas \ros~\cite{ros2} is now largely unavoidable in robotics,
successfully implementing distributed computation, standardized messaging and
packaging. We will see that \ros provides most -- but not all -- of the solutions we need.

This paper describes the software architecture required to support the \MST
project: enabling modularity across form, function, and location. Our system
must scale in both hardware and network dimensions, supporting reconfigurable
robot assemblies as well as the interplanetary communication patterns
envisioned in \cref{fig:interplanet}. Furthermore, this being at the laboratory
experiment stage, other requirements take precedence over space consideration.
Notably development, prototyping, developer onboarding should be simple, and
the architecture should be robust to frequent changes. A keen reader will
notice that this is also analogous to modularity.

\section{Method}
\subsection{Design Philosophy}
\label{sec:philo}

As highlighted by Conway~\cite{Conway1967HOWDC}, product and software
architecture converges towards the structure of the group of people making it.
In robotics, we find that the software is highly correlated to the shape of the
robot.

For us, this creates a critical contradiction, because a deeply modular robot
should not assume a shape. Hence, the developer must actively fight
against this pull toward a fix architecture and inappropriate assumptions.

To avoid such shortcomings, we propose a strong \textit{component-based software}
approach~\cite{fundamentals_of_prog}. We will avoid encoding relationships
between software components, because this would imply relationship between people
and robot shape. Instead, we treat the system as assemblies of loosely-coupled
components, where behavior emerges from data.

Core to this model is \textit{separation of concerns}~\cite{programming2007,
fundamentals_of_prog}. Components must not be aware of the internal logic -- or
even existence -- of other components. They should operate solely on resources,
exposed via a medium of data. Our architecture should be shaped by data, not by
the shape of the robot. It is then, the responsibility of higher-order
components -- solvers, planners, orchestrators, and AI agents -- to compose
behavior by acting on those data resources.

\subsection{Network}
\label{sec:meth_net}

As in industrial IoT (IIoT) systems, the network is a critical enabler of
modularity~\cite{iiot, zenoh, decentra_robot_mod}. Given our goal \cref{fig:interplanet} and philosophy \cref{sec:philo}, we require a data
transport layer offering strong separation of concerns, while supporting
long range and unreliable communications.

Traditional Data Delivery Systems (DDS) are efficient and relatively simple to
deploy~\cite{DDS}. However, they are strictly peer-to-peer, thus violate
separation of concerns by tightly coupling data flow between all participants.

In contrast, MQTT communicates through a central broker~\cite{MQTT}. While this decouples participants, it
introduces a single point of latency, failure, and bandwidth constraint.

Zenoh~\cite{zenoh} offers the best of both: peer-to-peer, brokered, and mesh
communication can coexist. Hence, the network topology can adapt to the mission
and robot. Independent software routers manage routing and optimization, making
network orchestration the sole responsibility of Zenoh.

\Cref{fig:interplanet} illustrates powerful concepts. Wireless
communication links are routed through Zenoh routers, isolating fragile and
long-range connections. Meanwhile, high-performance peer-to-peer links are used
on intra-base and intra-robot communication. Redundancy naturally emerges
through multiple radio paths and fallback wired links within each robot
assembly (e.g., between \textit{limb \#1, \#2} and \textit{wheel \#1, \#2}).

Due to hardware constraints, wired communication between modules is not yet available. All inter-module traffic is currently routed over Wi-Fi. Although a brittle, worst-case scenario, this setup is common in many robotic systems.

\subsection{Launch and Orchestration}

\label{sec:meth:orch}

As the number of robot modules grows -- each potentially composed of multiple
onboard computers -- the task of launching the correct processes, with
appropriate configuration, on the correct machines becomes exponentially more
complex, time-consuming, and error-prone.
Our orchestrator~\cite{fundamentals_of_prog} addresses this by reducing operator
intervention down to three actions per robot: update, build, and
start. While not fully automated, this already provides critical reliability
and speed during deployment.

Additionally, orchestration encodes static information that is not transmitted over the Zenoh network. Encoding every behavior as runtime messages would lead to unnecessary complexity and fragility.

Our orchestrator consists of the following steps:

\begin{enumerate}
    \item \textbf{Version Control}: Updates numerous private
        repositories to the correct branches for the current experiment.

    \item \textbf{Updater and Build}: Resolves dependencies, then uses the \textit{Doit} build system~\cite{doit} to resolve a graph of offline installation and compilation tasks.

    \item \textbf{Robot Assembler}: A \PY system that defines the software
        components of every module for use in a complete assembly. It deduces
        assembly-specific configurations (e.g., kinematic chains, URDF),
        computer-specific settings (e.g., executable paths, joints under
        control), and hardware-specific parameters (e.g., calibration data,
        motor type).

    \item \textbf{Component Launcher}: Identifies the host machine, launching
        the appropriate components. Execution is remotely started
        and internally delegated to the \ros launch system.

\end{enumerate}

\subsection{Components}
\label{sec:met:comp}

\begin{figure}[tb]
	\centering
	\includegraphics[height=0.3\columnwidth, trim={0 0 0 0}, clip]{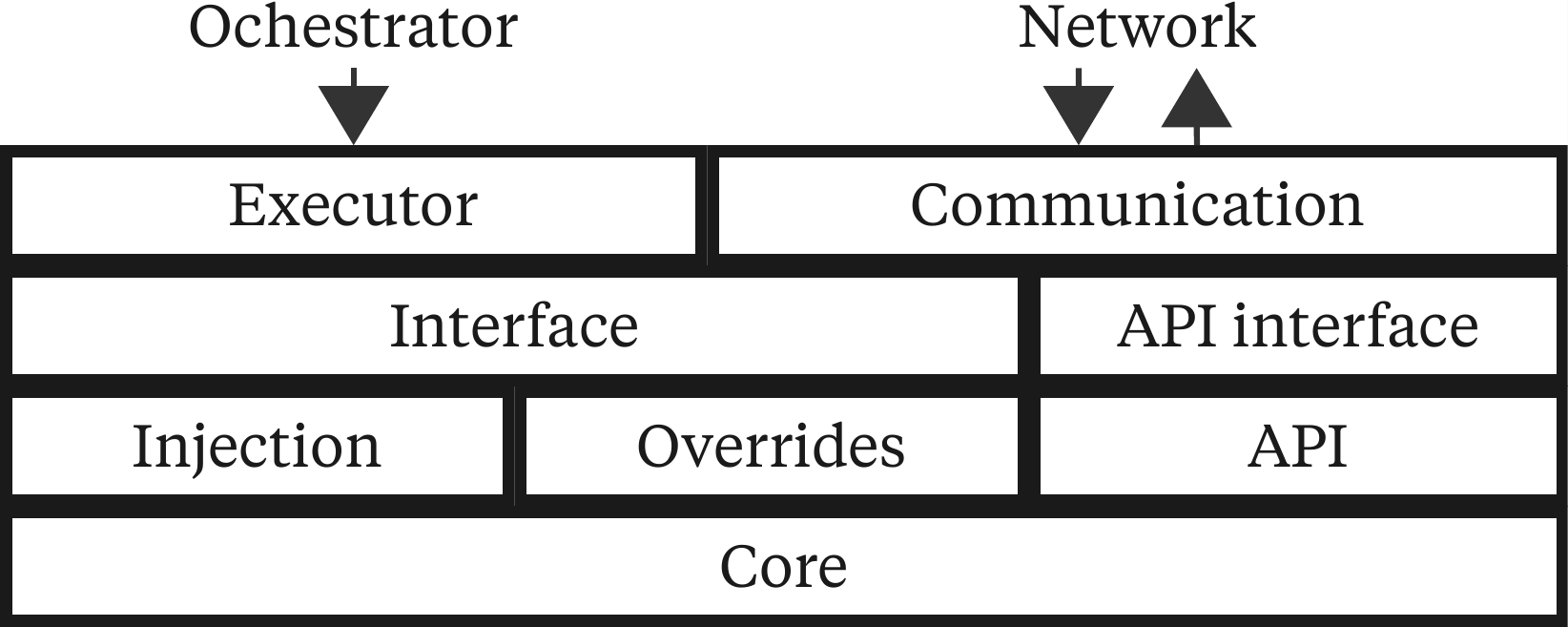}
    \caption{Structure of a software component, pushing for separation of concern down to the subcomponent level.}
	\label{fig:comp_struc}
\end{figure}

Our component structure, shown in \cref{fig:comp_struc} and detailed in \cref{tab:component_blocks}, is designed to minimize robot-specific code and enforce modularity down to the subcomponent level.

\begin{table}[tb]
\vspace{1.8mm}
\centering
    \caption{Roles and scopes of subcomponents \cref{fig:comp_struc}.}
\label{tab:component_blocks}
\begin{tabular}{rlll}
    \textbf{Block} & \textbf{Role} & \textbf{Specific to} & \textbf{Notes} \\
    \hline
    Core          & Main logic & None & Pure python\\
    Injection     & Extend the core & Core type & Tested, tunable \\
    Overrides     & Alter the core & Robot &  System-specific\\
    API           & Simple interface & None            & Beginner friendly \\
    Interface     & Bridge & Core type    & \ros wrapper \\
    Executor      & Call the core & OS     & \ros executor \\
    Comm. & Data Transport & Architecture     & \ros over  Zenoh\\
\end{tabular}
\end{table}

Modular robotics present numerous hardware and system-specific
variations. To accommodate each specificity, the code often branches with multiple execution paths, ultimately becoming difficult to maintain.

Our model avoids this by isolating robot-specific logic within the
well-defined \texttt{Injection} and \texttt{Override}
blocks~\cite{programming2007,fundamentals_of_prog} -- both of which are
chosen by the orchestrator (\cref{sec:meth:orch}). For example,
\texttt{Injection} may be a telemetry system, while
\texttt{Override} handles a broken joint.

We deliberately avoid using launch parameters for hardware-specific options.
First, \texttt{Core} developers cannot anticipate all possible customization
points. Second, launch parameters are limiting and static, thus unsuitable for
runtime behavior. Third, following \textit{separation of concerns}, the
launcher should only expect to interact with the original \texttt{Core} class.
While tightly integrated, the orchestrator aims to be an independent system.

The highly reusable \texttt{API} subcomponent abstracts low-level communication
and safety-critical logic behind a consistent interface. As shown in our
previous work~\cite{ms_api}, actuator synchronization can be robustly
implemented without requiring detailed system knowledge -- allowing safe
operations on any actuators.

The strength of this whole structure lies in its composability: each
subcomponent can be reused, swapped, or tuned with minimal branching and coupling to other systems.

\subsection{Implementation and Test Campaigns}

\label{sec:implem}

The system was implemented on the \MBs (\cref{fig:mainx4}) and PCs of
collaborators interacting with the robots. \ros serves as our backbone, providing
both DDS and Zenoh compatibility. The \MBs and user PCs run Ubuntu 22.04 or
24.04 depending on hardware compatibility. Although \ros does not fully meet
our \textit{separation of concerns} requirement -- due to its tight coupling
between operating system of all participants -- the need for rapid prototyping
and onboarding motivated its adoption.

The system was tested and improved gradually by intensive test campaigns: three
field tests totaling nine weeks at the JAXA sand field~\cite{jaxafield}; a
five-day experiment in DLR’s LUNA lunar simulation facility
(\cref{fig:mainx4}a)~\cite{lunafield}; and a six-day campaign at the Osaka 2025
Universal Exposition, with daily operations of ten hours. Referring to
\cref{fig:interplanet}, these tests implemented MoonBase~\#1, where multiple
students simultaneously operated multiple robots locally. Some tests also
involved remote operations, simulating a lunar base controlling robots from
afar. In one proof-of-concept trial at the Osaka Exposition, we demonstrated
precise teleoperation across multiple wireless links: Operator $\rightarrow$
WiFi $\rightarrow$ Internet $\rightarrow$ Cellular 5G $\rightarrow$ WiFi
$\rightarrow$ Robots.

Our primary goal in those campaign was to provide a flexible platform for all
our collaborators to control the robots. Those experiments provided us with
critical feedback, observation and data (rosbag), shaping the current system
architecture.

\section{Results}

\label{sec:result}

\subsection{Modularity}
\label{sec:res:mod}

\begin{figure}[tb]
    \vspace{2.8mm}
    \centering
    \begin{overpic}[width=1\columnwidth, trim={500 600 0 300}, clip]{"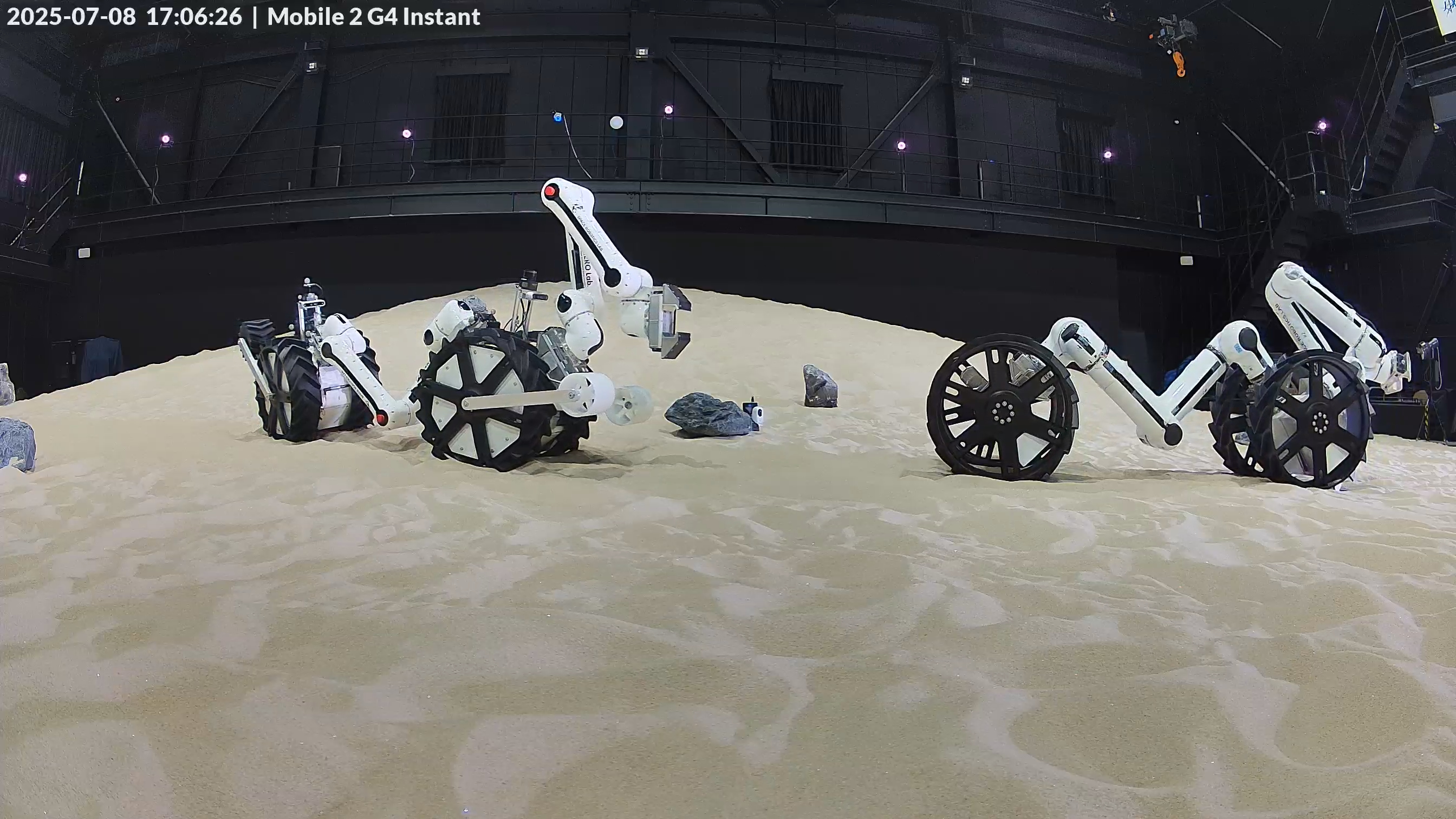"}
        \put(2,23){\color{white}\textbf{(a)}}
    \end{overpic}
    \par\vskip3pt
    \begin{overpic}[width=1\columnwidth, trim={300 100 200 200}, clip]{"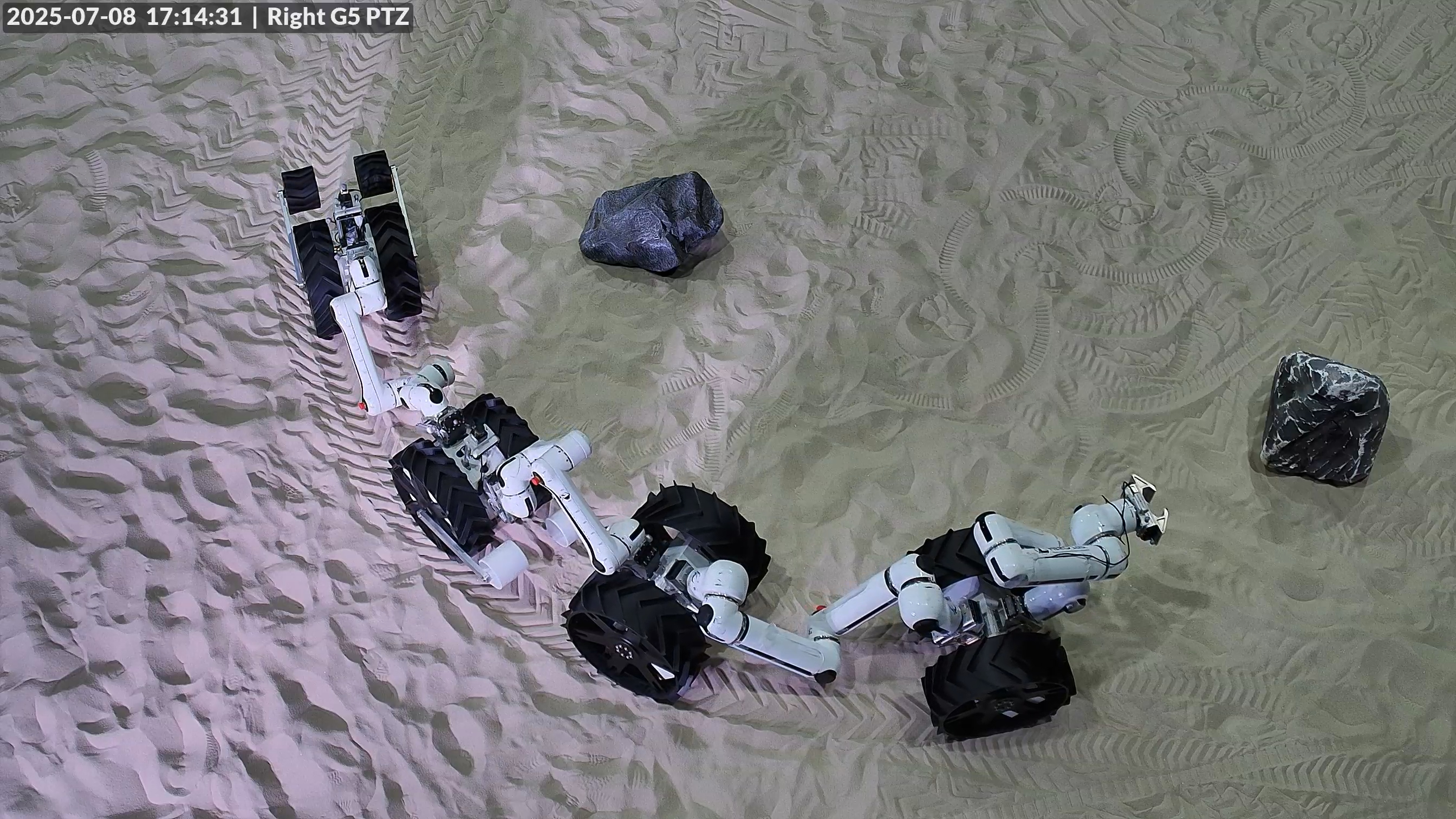"}
        \put(2,50){\color{white}\textbf{(b)}}
    \end{overpic}
    \caption{Two assemblies of independent robot modules, connecting to form
    one large assembly. (a) \text{Left}: Moonbot Dragon assembly, made of 2 H-line wheels V2 and 2 H-line limbs V2. \text{Right}: Variant with V1 modules. (b) The large heterogeneous assembly moving.}
    \label{fig:train}
\end{figure}

Our system achieves broad modularity. We curretnly support seven robot modules across two hardware families (H-line and G-line). Importantly, those family lines have major hardware differences -- number of degree-of-freedom, motor controller, embedded computer -- reflected in software by changes to the \texttt{Injection} and \texttt{Override} blocks \cref{fig:comp_struc}. This diversity shown \cref{fig:mainx4,fig:train} is comprehensively managed by the orchestrator, allowing for mixed configurations.

As outlined in \cref{sec:intro}, modularity must extend beyond hardware to the full robotic system. During experiments involving two Dragon assemblies (\cref{fig:train}.a) multiple ground control teams operated the robots concurrently from separate machines. \Cref{tab:flex} details the 60 components simultaneously involved. Each robot and control module acted independently, communicating solely via data -- enabling decentralized operation with minimal coupling.

Operator collaboration emerged naturally, with seamless task sharing across independent component. In \cref{fig:train}, coordinated steering was achieved as two human operators controlled the front and rear halves -- demonstrating the system’s decentralized and composable behavior in practice.

\begin{table}[tb]
    \vspace{1.8mm}
\centering
    \caption{Components during the distributed operation of the Dragon assemblies in \cref{fig:train}. (GC: Ground Control)}
\label{tab:flex}
\begin{tabular}{rcl}
    \textbf{Component} & \textbf{Count} & \textbf{Location} \\
    \hline
    Motor interface & 8  & Robot\\
    Joint Manager & 8  & Robot\\
    Kinematics Manager & 4  & Robot\\
    Health monitor & 8  & Robot\\
    Location publisher & 16  & Robot\\
    Human operator & 3  & GC\\
    Autonomous operator & 6  & GC\\
    Data monitoring & 5  & GC\\
    Data archiving & 2  & GC\\
\end{tabular}
\end{table}

\subsection{Development Time}
\label{sec:res:dev}

\begin{table}[tb]
\vspace{1.8mm}
\centering
\caption{Estimated time required for common development and deployment tasks under our system.}
\label{tab:time}
\begin{tabular}{rlll}
    \textbf{Action} & \textbf{Time estimate} \\
    \hline
    Adding a new robot family line          & 1-2 week\\
    Adding a new module to a family         & 2 days\\
    Adding an updated module revision       & 1 day\\
    Coding a new assembly template          & 1 day\\
    Swapping robots of an assembly          & 20 seconds\\
    Getting an assembly's parameters \& URDF          & Instantaneous\\
    \hline
    Entire software instalation & 30 minute\\
    Switching between sim and real & Instantaneous\\
    Starting a remote robot & 1 minute\\
    \hline
    Learning \ros & 2 weeks\\
    Onboarding a new developer         & 1 week\\
    Learning only the \PY API         & 2 day\\
    \hline
    Starting one Zenoh router & 1 minute\\
    Applying communication layer changes & 1 minute\\
\end{tabular}
\end{table}

The component-based architecture substantially reduced development effort by focussing on modularity. This proved critical during field-testing where experiments are conducted by many students and robots often replaced. The estimated durations of the common operations are outlined \cref{tab:time}.

\textbf{System Integration.} Adding or modifying robot families, modules, or assemblies requires minimal changes -- mostly confined to the \texttt{Injection}, \texttt{Override}, and metadata layers. The orchestrator automates system configuration and URDF generation, enabling rapid prototyping and reuse.

\textbf{Deployment and Simulation.} Setting up a full robot environment is fast and reproducible with the orchestrator. Simulation is deeply integrated: hardware components are automatically replaced with virtual counterparts when orchestration detects a non-robot host. This makes sim-to-real transitions instantaneous and consistent.

\textbf{Onboarding and Productivity.} The \PY \texttt{API} dramatically lowers the barrier to entry. Developers can interact with components without needing to understand low-level \ros concepts, which reduces onboarding time and did prevent many hardware damages.

\textbf{Communication Reconfiguration.} The separation of communication logic enables fast, non-intrusive changes to network transport. These changes can be deployed quickly -- critical during large experiments involving many end-points.

\subsection{\ros Limitations}

\label{sec:res:ros}

While \ros is a backbone of our system, its node-centric design conflicts with
our data-oriented model. The \ros ecosystem encourages minimizing the number of
nodes and data endpoints (e.g., publishers and subscribers). Additionally, all
endpoints contribute to a single, globally shared ``rosgraph''. This tightly
couples components and breaks separation of concerns.

This is especially evident in the handling of transforms: all transforms in
\ros are traditionally aggregated on a global \texttt{/tf} topic. As a result,
an Earth station that only needs the pose of a single robot must listen to the
entire -- interplanetary -- transform stream. Data aggregation, batching,
routing and stream optimization should instead be delegated to the network, by
keeping data simple and decoupled.

Beyond architectural concerns, this introduces excessive network traffic,
processing, and fragility. These limitations are not only theoretical, as will
demonstrate the next section.

Finally, the incompatibilities between \ros versions -- tied to very specific
and heavy distributions of Ubuntu -- is a major drawback for large scale
collaboration. Teams, components and robots become incompatible and fragmented
because of requirements unrelated to the project.

\subsection{Network}
\label{sec:res:net}

\begin{table}[tb]
\vspace{1.8mm}
\centering
\caption{Comparison of network behavior between DDS and Zenoh, measured on one robot with 6 other active robots.}
\label{tab:net_prob}
\begin{tabular}{rlll}
    &\textbf{DDS} & \textbf{Zenoh} \\
    \hline
    Component startup & 10 s to 1 min & 1 s\\
    Startup bandwidth & 69 Mb/s & 1.65 Mb/s\\
    Startup stutter & 20 s & 1s\\
    \hline
    Average latency & 1.6 ms & 2.2 ms\\
    Average transmitted bandwidth & 1,200 Kb/s & 1,130 Kb/s\\
    Average received bandwidth & 550 Kb/s & 250 Kb/s\\
    Peak bandwidth & 69 Mb/s & 2 Mb/s\\
    Connectivity recovery time & 20s & 2s\\
    Max stable robot count & 4 & 10+\\
\end{tabular}
\end{table}

We began development using \ros's default DDS backend, improved by 
discovery servers. However, DDS quickly suffered from severe
startup and recovery overheads. The analysis of field test's data (rosbags)
highlights those problems \cref{tab:net_prob}. Most problematic being that not
only \ros, but network-wide stutters happen using DDS.

These issues persisted despite using a high-end Wi-Fi 7 setup (multi-link operation, dedicated APs, isolated traffic) -- conditions more favorable than most real-world deployments. Still, DDS was not reliable enough to support more than four robots transmitting only basic joint and state information.

As anticipated in \cref{sec:meth_net}, DDS’s peer-to-peer design fails to scale under our distributed, modular architecture.


Switching to Zenoh as \ros communication middleware resolved these issues. The clean separation in our architecture (\cref{sec:meth:orch,fig:comp_struc}) allowed us to migrate within a day. With Zenoh, reactiveness improved, bandwidth stabilized, and collaboration across many components (\cref{tab:flex}) became seamless -- allowing for reliable, large assemblies (\cref{fig:train}). Zenoh ultimately allowed seamless remote operations, notably the Osaka trial previously explained \cref{sec:implem}.

The startup stutter is however still slightly present with Zenoh, but it is short and only affects \ros communications.

\section{Discussion}

\subsection{Future Transport Layer Direction}

While Zenoh is already our \ros communication middleware, its full benefits are constrained by \ros -- as preempted by \cref{sec:implem}. A full transition to pure Zenoh would resolve many architecture problems highlighted \cref{sec:res:ros}:

\begin{itemize}
    \item Only data flows, reducing overhead and brittleness.
    \item Lightweight deployment: Low system requirements.
    \item Scalability: Thousands of data sources and sinks.
    \item Advanced routing: data is precisely managed and monitored by the routers.
    \item Storage: Native support for distributed data storage (e.g., telemetry, LiDAR scans, video logs).
\end{itemize}

We are actively exploring a full Zenoh implementation to better match our data-oriented, distributed model. However, this comes with ecosystem tradeoffs: \ros tools would need reimplementation, collaborators would need to learn new skills, and compatibility with the broader \ros community would be lost. As such, our near-term goal is a pure Zenoh prototype, then a compatibility bridge to \ros.

Finally, future work will include testing under degraded conditions resembling space networks -- high-latency, low-bandwidth, and intermittent links.

\subsection{Toward Smarter Orchestration}

Currently, the orchestrator relies on a manual launch procedure: an operator specifies the assembly and initiates the launch. While manual, it is infrequent and human safety assessment is essential -- especially in high-stakes space robotics. Furthermore, the assembly information takes less than 20 seconds to specify via the robot-assembler detailed \cref{sec:meth:orch,tab:time}.

We believe that full automation of the orchestrator offers limited practical value. Automatically detecting assemblies would require each robot to identify itself, introducing coupling and added complexity.

A more beneficial improvement would be enabling local orchestrators to synchronize with a central database. Currently, they operate fully offline and are manually updated on each robot -- which is error-prone and time-consuming. Network-based synchronization would improve maintainability without compromising modularity or safety.

\subsection{Applicability}
\label{sec:dis:app}

This system has not yet been tested under space-like constraints. Our hardware uses standard consumer components, and our communication relies on terrestrial cellular network, Wi-Fi and LAN, rather than space-qualified radios. This is by design: the primary objective of this work is not a space-specific solution, but to discover architectural guidelines applicable to future modular and distributed robotic systems.

Our goal is to surface principles that scale across all domains through our prototypes. The lessons presented throughout the paper -- component separation (\cref{sec:philo,sec:met:comp,sec:res:mod}), data-centric communication (\cref{sec:meth_net,sec:res:net}), and orchestrated deployment (\cref{sec:meth:orch}) -- hold beyond space applications.

Furthermore, our most fundamental components -- joint manager, kinematic manager, API \cite{ms_api}, operator interface -- are available open-source in our \MS software (\url{https://github.com/2lian/Motion-Stack}).

Ultimately, we advocate for software structures that are resilient to change: of robot shape, team composition, communication infrastructure, or mission scope.

\section{Conclusion}

This work presents a software architecture and deployment strategy for
distributed, heterogeneous modularity in robotics. Building on principles of
separation of concerns, component-based design, and data-centric communication,
we demonstrated a system capable of operating large, reconfigurable assemblies
composed of independently designed and controlled modules.

The modularity, flexibility and network functionality of our architecture was improved and validated through extensive field-testing, while maintaining low
development overhead. Key enablers include the
orchestrator (\cref{sec:meth:orch}), which automates deployment across diverse
modules; the clean component structure (\cref{sec:met:comp}); and the
Zenoh-based communication backbone (\cref{sec:meth_net}), which resolves
limitations inherent in traditional \ros and DDS systems
(\cref{sec:res:ros,sec:res:net}).

While not yet space-hardened, our system was deliberately designed with
generality in mind (\cref{sec:dis:app}). Our architecture aims to be agnostic
to physical implementations, and thus applicable to industrial, terrestrial,
and interplanetary deployments alike. Future work will address space-specific
constraints, further promising Zenoh tooling, and improving orchestrator via
centralized server. This contribution serves as a practical reference for
designing scalable, modular robotic systems. 

More broadly, we advocate for adaptive, distributed deployments as a foundation
not just for modular space robotics, but for the future of robotics as a whole.

\bibliography{bib.bib}

\end{document}